\documentclass{article}

\usepackage[preprint]{neurips_2022}

\usepackage[utf8]{inputenc} 
\usepackage[T1]{fontenc}    
\usepackage{hyperref}       
\usepackage{url}            
\usepackage{booktabs}       
\usepackage{amsfonts}       
\usepackage{nicefrac}       
\usepackage{microtype}      
\usepackage{xcolor}         
\usepackage{tikz}
\usepackage{amsfonts}
\usepackage{amsmath}
\usepackage{physics}
\usepackage{bm}
\usepackage{float}
\usepackage{cleveref}
\usepackage{mathtools}
\usepackage{caption}
\usepackage{subcaption}

\usepackage{thmtools}
\usepackage{thm-restate}

\newcommand{\E}[0]{\mathbb{E}}

\newcommand{\Var}[0]{\mathrm{Var}}

\newcommand\ci{\perp\!\!\!\perp}

\newtheorem{lem}{Lemma}
\newtheorem{assumption}{Assumption}
\newenvironment{prf}{\paragraph{Proof:}}{\hfill$\square$}

\crefname{assumption}{Assumption}{Assumptions}
\crefname{cor}{Corollary}{Corollaries}
\crefname{lem}{Lemma}{Lemmas}
\crefname{dfn}{Definition}{Definitions}

\DeclarePairedDelimiterX{\infdivx}[2]{(}{)}{%
  #1\;\delimsize\|\;#2%
}
\newcommand{\infdiv}{D\infdivx}

\title{Off-Policy Evaluation for Embedded Spaces}

\author{Jaron J.R. Lee\\
Adobe Research \\
\texttt{jaron2005@gmail.com} \\
\And
David Arbour \\
Adobe Research\\%
\texttt{arbour@adobe.com} \\
\And
Georgios Theocharous \\
Adobe Research \\
\texttt{theochar@adobe.com}
}

\begin{document}

\maketitle

\begin{abstract}
 Off-policy evaluation methods are important in recommendation systems and search engines, where data collected under an existing logging policy is used to estimate the performance of a new proposed policy. A common approach to this problem is weighting, where data is weighted by a density ratio between the probability of actions given contexts in the target and logged policies. In practice, two issues often arise. First, many problems have very large action spaces and we may not observe rewards for most actions, and so in finite samples we may encounter a positivity violation. Second, many recommendation systems are not probabilistic and so having access to logging and target policy densities may not be feasible. To address these issues, we introduce the featurized embedded permutation weighting estimator. The estimator computes the density ratio in an action embedding space, which reduces the possibility of positivity violations. The density ratio is computed leveraging recent advances in normalizing flows and density ratio estimation as a classification problem, in order to obtain estimates which are feasible in practice.
\end{abstract}

\section{Introduction}
Off-policy evaluation (OPE) has become increasingly important in many areas such as recommendation systems and search engines. A variety of methods, based off the idea of inverse propensity score weighting (IPW), have been developed to address these problems. These methods for OPE have been applied in areas such as advertising \citep{swaminathanOffpolicyEvaluationSlate2017}, music recommendation \citep{mcinerneyCounterfactualEvaluationSlate2020}, and search engine ranking \citep{liOfflineEvaluationRanking2018}.


However, applied practitioners attempting to implement these off-policy evaluation methods can encounter some difficulties. First, existing OPE methods require absolute continuity. However, in practice either this is violated, or the probability of observing a given action in a context is low enough that in finite samples and large enough action spaces it is not observed, leading to poor estimator performance. Second, 
OPE methods vary in what they assume of target and logging policies -- sometimes one or more of the policy density functions are available \citep{kallusPolicyEvaluationOptimization2018}, whereas other times it is assumed that only samples from the policies are available \citep{arbourPermutationWeighting2021,sondhiBalancedOffPolicyEvaluation2020}. In this paper we take the view that since practice recommendation systems are not always probabilistic systems, assuming access to density functions is not reasonable. However, dealing with estimation of density ratios is itself a challenging problem and naively tackling this will also lead to poor performance.

The goal of this paper is to improve the practical applicability of OPE methods by addressing the above concerns. To tackle the first concern, we introduce the idea of using action embeddings in estimating policy values. In many large action space problems, a given action may be quite similar to other actions for the purposes of the reward. Using a suitable embedding can therefore improve OPE methods since many actions might have the same embedding (and thus lead to better estimates of the weights). 

To tackle the second concern, we leverage ideas from the density ratio estimation literature - first, that we can recast the problem of inferring the importance sampling weights using density ratio estimation via probabilistic binary classification~ \cite{sondhiBalancedOffPolicyEvaluation2020}, and second, that density ratio estimation can be improved by an invertible transformation of the embedding space through the use of normalizing flows per \cite{choiFeaturizedDensityRatio2021}. Both these ideas improve the quality of estimating the weights when we do not have policy densities available.

The result is the featurized embedded permutation weighting estimator, which is a method for off-policy evaluation in normalizing-flow transformed action embedding spaces, where the density ratios are estimated using binary classification techniques. 



To summarize, this paper contributes the following:
\begin{itemize}
\item An estimator for this method that exploits previous work linking density ratio estimation and binary probabilistic classification, to estimate the density ratio of the target policy to the logging policy;
\item Theoretical guarantees on the bias and variance of this method, under random and deterministic embeddings;
\item Empirical validation of the method on synthetic and semi-synthetic experiments. 
\end{itemize}
The outline of this paper is as follows. In Section \ref{sec:standard-problem} we formulate the typical off-policy evaluation problem, and point out a common failure mode. In Section \ref{sec:approach} we present embedded off-policy estimators that exploit embedded spaces, which return results even in the presence of absolute continuity assumption violations. In Section \ref{sec:analysis} we present bounds on the bias and variance of one of the estimators, the embedded permutation weighting estimator, and show that the method exchanges the absolute continuity assumption for some additional bias and variance of the estimate. In Section \ref{sec:results} we present results on toy and realistic simulated data. 
We conclude with Section \ref{sec:conc}.
\subsection{Related work}
The use of inverse probability weights goes back to \cite{horvitzGeneralizationSamplingReplacement1952}. The technique has become common in the off-policy evaluation literature \cite{liUnbiasedOfflineEvaluation2011,dudikDoublyRobustPolicy2014,thomasDataefficientOffpolicyPolicy2016}. However, as noted in \cite{sondhiBalancedOffPolicyEvaluation2020} these methods assume that the policy densities are known exactly, which is not always the case in practice.

This work relies on density ratio estimation techniques based on binary probabilistic classification, which dates to at least \cite{qinInferencesCasecontrolSemiparametric1998}.  More immediately, we leverage results from \cite{sondhiBalancedOffPolicyEvaluation2020} who considered the problem in the off-policy evaluation context. \cite{menonLinkingLossesDensity2016} considered the problem of linking density ratio estimation to probabilistic classification.
Non-classification based density ratio methods otherwise rely on kernel methods  \citep{huangCorrectingSampleSelection2006,sugiyamaDensityRatioEstimation2012}

Normalizing flows are a deep neural network method by which a given probability density of arbitrary complexity can be transformed into a simpler one using an invertible transformation. This is done by carefully constructing the neural network such that it is composed only of invertible transformations, and has been of great interest \citep{kingmaGlowGenerativeFlow2018,dinhDensityEstimationUsing2016}. Recently, these techniques were applied to the problem of density ratio estimation \cite{choiFeaturizedDensityRatio2021}, which proposed fitting a normalizing flow over the combined data from both densities of the ratio, before applying the density ratio estimation technique of choice.

The authors are aware of an independent and contemporaneous submission \citep{saitoOffPolicyEvaluationLarge2022}) which also leverages embedded spaces for off-policy evaluation. By contrast, this paper focuses on practical estimation techniques -- exploring the permutation weighting method applied to this problem in greater detail, using normalizing flows for improving density ratio estimation, and issues surrounding deterministic embedding systems.

\section{Problem Setting}\label{sec:standard-problem}
The problem considered in this paper is as followed. 
Let $C \in {\cal C}$ denote the context, $\mathcal{A} \in {\cal A} =  \{1, \ldots, d\}$ a set of possible discrete actions, and $Y \in {\cal Y}$ to be a scalar such that $Y \sim p(Y \mid A, C)$. 
Finally, let $\pi_0(A \mid C)$ be the observed logging policy, and $\pi_1(A \mid C)$ be the target policy.

The distribution of data under the logging policy is $p_0 = p_0(Y, A, C) = p(Y \mid A, C)\pi_0(A \mid C) p(C)$, and under the target policy it is $p_1 = p_1(Y, A, C) = p(Y \mid A, C)\pi_1(A \mid C) p(C)$.

We observe data as tuples $(C, A, Y)_{i=1}^N$, generated under logging policy $p_0(Y, A, C) $. We denote the set of observed actions under the logging policy $\pi_0$ as $\mathcal{A}_0$ with  $d_0 = \abs{\mathcal{A}_0}$ and likewise for the target policy $\pi_1$.


The goal is to estimate the expected value of rewards under the target policy, 
\begin{equation} \label{eqn:policy_value}
V(\pi) = \E_C [\E_{\pi_1} [\E[Y \mid A, C]]]
\end{equation}

To do this, standard assumptions are made \citep{hernanCausalInferenceWhat2020}.

\begin{assumption} \label{ass:consistency}
Consistency: $Y(a)$ is equal to $Y$ when $A=a$
\end{assumption}
\begin{assumption} \label{ass:conditional_ig}
Conditional ignorability : $Y(a) \ci A \mid C$
\end{assumption}
\begin{assumption}\label{ass:action_positivity}
Absolute continuity in actions: Whenever $\pi_1 (A \mid C) >0 $ then $\pi_0 (A \mid C) >0$. 
\end{assumption}

\cref{ass:consistency} implies that the causal mechanisms are stable under intervention. \cref{ass:conditional_ig} implies that adjusting for the context $C$ is enough to block all confounding between the action and the potential reward. \cref{ass:action_positivity} ensures that the density ratio is well-defined.

Then, the inverse probability weighting (IPW) estimator is
\[\hat{V}_{IPW} = \frac{1}{N} \sum_{i=1}^N Y_i \frac{\pi_1(A_i \mid C_i)}{\pi_0(A_i \mid C_i)}\]

To stabilize weights, we can instead divide by the sum of the weights. This leads to the IPWS estimator which is
\[\hat{V}_{IPWS} = \frac{\frac{1}{N} \sum_{i=1}^N Y_i \frac{\pi_1(A_i \mid C_i)}{\pi_0(A_i \mid C_i)}}{\frac{1}{N} \sum_{i=1}^N  \frac{\pi_1(A_i \mid C_i)}{\pi_0(A_i \mid C_i)}} \]
and is an unbiased estimator with variance at least as small as that of $\hat{V}_{IPW}$ \cite{hiranoEfficientEstimationAverage2003}.

If $\mathcal{A}_0 \neq \mathcal{A}_1$ then absolute continuity is violated, and previously unseen actions in the target policy cannot be handled. This greatly limits the kinds of target policies that can be evaluated in practice, if new actions are introduced.

The direct method estimator (also known as the g-formula estimator) aims to estimate a model for the reward given the action and context. It is given as 
\begin{equation} 
\hat{V}_{DM} = \frac{1}{N} \sum_{i=1}^N \pi_1(A_i \mid C_i) \hat{\E}[Y \mid A_i, C_i] \label{eqn:dm}
\end{equation}
 for $\hat{\E}$ a regression model. 

In the case where we do not have the policy densities available, we can first estimate models $\hat{\pi}_0, \hat{\pi}_1$ using a suitable model for the logging and target policies before estimating the IPW/DM method. We distinguish these versions of these estimators with an asterisk:
\[\hat{V}_{DM^*}= \frac{1}{N}\sum_{i=1}^N \hat{\pi}_1(A_i \mid C_i) \hat{E}[Y \mid A_i, C_i], \quad \hat{V}_{IPW^*}=\frac{1}{N} \sum_{i=1}^N  Y_i\frac{\hat{\pi}_1 (A_i \mid C_i)}{\hat{\pi}_0 (A_i \mid C_i)} \]

\section{Embedded Off-Policy Evaluation} \label{sec:approach}
In some applications, such as image recommendation, many actions may be similar in some sense, and it is reasonable to expect similar responses to these. However, standard OPE approaches would treat these actions and contexts as distinct, and thereby fail.

Conversely, developments in other areas of machine learning have led to great advances in image and word embedding systems. These systems can provide a vector encoding of the salient aspects of their inputs, which we would expect to group together similar actions. Examples of such systems include \texttt{word2vec} \citep{mikolovDistributedRepresentationsWords2013} or \texttt{CLIP} \citep{radford2021learning}.

Therefore, we assume the following: 
\begin{assumption}\label{ass:map}
Embedding: we have access to a map such that for each $A\in \mathcal{A}$, we have a random embedding $G$
such that 
\[p(Y A, G, C)= p(Y \mid G, C) p(G \mid A) \pi (A \mid C) p(C)\]
\end{assumption}
This assumption ensures that  the policy value in \cref{eqn:policy_value} is not disturbed under the embedding, since the embedding captures all relevant information of the action for determining the reward.



In place of \cref{ass:action_positivity}, we instead assume the following:
\begin{assumption} \label{ass:positivity}
Absolute continuity in embedding space: $\pi_1(G \mid C) > 0 \implies \pi_0(G \mid C) > 0$
\end{assumption}

Notice that \cref{ass:positivity} only requires absolute continuity over the embedding space rather than the original space. By \cref{ass:map}, $\pi(G \mid C) = \sum_A p(G \mid A) \pi(A \mid C)$. If the embedding $p(G \mid A)$ has positive support for all $G, A$ then $p(G \mid C)$ is guaranteed positive support, no matter the distribution of $p(A \mid C)$. However, no such guarantee holds if $p( G \mid A)$ does not have positive support (e.g. if it is a deterministic map).

\subsection{Estimation when both policy densities are available}
In the first instance, assume we have the target and logging densities $\pi_1, \pi_0$. In practice, this might happen if the target and logging policies happened to be modelled by a probabilistic model. Then we can directly evaluate the following estimator under data from the logging policy $(Y_i, A_i, G_i, C_i)_{i=1}^N$:
\[\hat{V}_{EIPW} = \frac{1}{N} \sum_{i=1}^N Y_i \frac{\pi_1(G_i \mid C_i)}{\pi_0(G_i \mid C_i)}\]
where $\pi(G \mid C) = \sum_A p(G \mid A) p(A \mid C)$.

The embedded direct method estimator (also known as the g-formula estimator) aims to estimate a model for the reward given the embedded action and context. It is given as 
\begin{equation}
\hat{V}_{EDM} = \frac{1}{N} \sum_{i=1}^N \pi_1(G_i \mid C_i) \hat{\E}[Y \mid G_i, C_i] \label{eqn:embed_g}
\end{equation}
 for $\hat{\E}$ a regression model fitted on data $(Y_i, A_i, G_i, C_i )_{i=1}^N$ from the logging distribution.

\subsection{Estimation when one or more policy densities are not available}
In practice, most policies are modelled by machine learning systems which may be non-probabilistic classifiers or regressors, meaning that we are able to sample actions given a context from a policy, but not evaluate the probability density of doing so. We follow the approach of \cite{sondhiBalancedOffPolicyEvaluation2020} in estimating density ratios using binary classification.

Let samples from the logging policy be $(Y_i, A_i, G_i, C_i)_{i=1}^N$ and the target policy $(A'_i, G_i', C_i)_{i=1}^N$.
From these samples, we create a dataset ${\cal D} = [(G_i, C_i)_{i=1}^N, (G_j', C_j)_{j=1}^N]$ formed by concatenating embeddings and contexts from the logging and target policy data. We define a variable $Z$ which takes value $0$ if row $i$ of ${\cal D}$ was drawn from the logging dataset, and $1$ otherwise. We fit a probabilistic binary classifier with label $Z$ against features  $G ,C$. The classifier recovers $\eta(G ,C)=p(Z=1 \mid G, C)$, with estimate $\hat{\eta}$. Then, 
\begin{align*}
    w(G, C) = &\frac{\eta(G, C)}{1 - \eta(G ,C)} =\frac{p(Z=1 \mid G ,C) }{p(Z=0 \mid G , C)} = \frac{p(G, C \mid Z=1) p(Z=1) p(G , C)}{p(G , C \mid Z=0) p(Z=0) p(G, C)}, \\
    &= \frac{\pi_1(G \mid C)p(C)}{\pi_0(G \mid C)p(C)}= \frac{\pi_1(G \mid C)}{\pi_0(G \mid C)} 
\end{align*}

by application of Bayes' rule and standard probability operations.

The implication of this result is that the ratio of the densities can be estimated by fitting a probabilistic classifier $\hat{\eta}$, which we denote $\hat{w}(G ,C) = \hat{\eta}(G ,C) / (1 - \hat{\eta}(G ,C))$
This results in the \emph{embedding permutation weighting} estimator: 
\begin{equation}
\hat{V}_{EPW} = \frac{1}{N} \sum_{i=1}^N Y_i \hat{w}(G_i,C_i)  \label{eqn:embedding-ope-no-kern}
\end{equation}
Empirically, \cite{sondhiBalancedOffPolicyEvaluation2020} noted two further tricks to improving performance. First, we can consider the method of \cite{kallusPolicyEvaluationOptimization2018} where kernel methods are used to improve performance. The intuition here is that we can leverage the sampled target policy by upweighting the logging policy outcomes whose actions are  closest to the target policy. Second, we can use self-normalizing estimator ideas to reduce variance \cite{hiranoEfficientEstimationAverage2003}. Combining these ideas produces the \emph{embedded permutation weighting self-normalized} estimator: 
\begin{equation}
\hat{V}_{EPWS} = \frac{\sum_{i=1}^N Y_i \hat{w}(G_i ,C_i) K\left(\frac{G'_i - G_i}{h}\right) }{\sum_{i=1}^N \hat{w}(G_i ,C_i) K\left(\frac{G'_i - G_i}{h}\right)}\label{eqn:embedding-ope}
\end{equation}

where $G_i, G'_i$ are drawn from the logging and target policies respectively, $K$ represents some kernel function (for example, the radial basis function kernel),  and $h$ is the bandwidth of that kernel selected in some reasonable fashion (e.g. $K$ being the radial basis function kernel with $h$ selected via median distance heuristic).

\subsection{Improving density ratio estimation using normalizing flows}

\begin{figure}[t]
  \centering
  \includegraphics[width=.5\textwidth]{"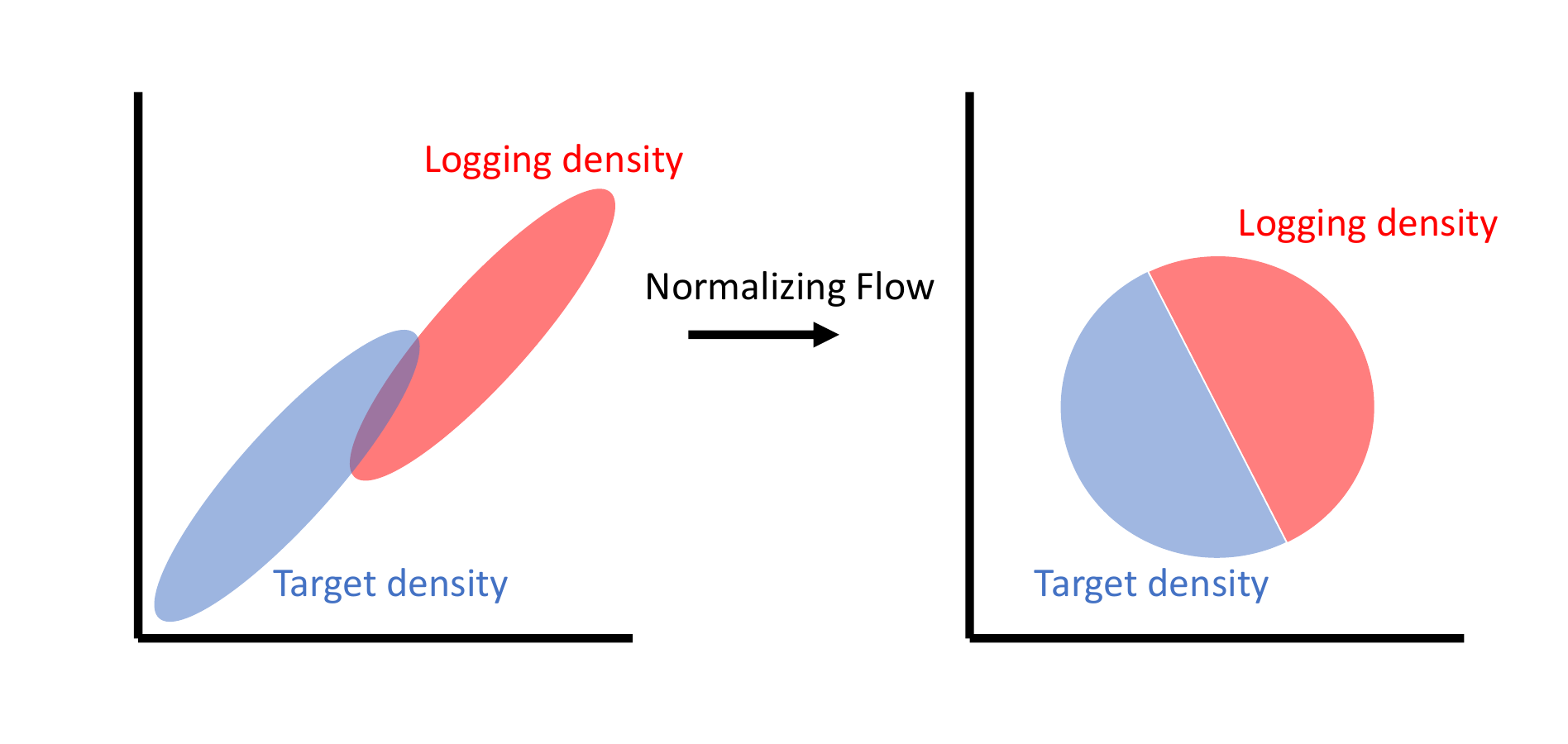"}
  \caption{A visual representation  of the effect of the normalizing flow, adapted from \cite{choiFeaturizedDensityRatio2021}. The logging and target densities (left) may be far apart in the embedding space, but after transformation  (right) they are reshaped into a standard Gaussian and thus brought closer together.}
\end{figure}

The EPWS estimator mitigates lack of overlap between the target and logging policies in the action space, but depending on the structure of the embedding itself there might be a considerable distance between the logging and target densities in this space. The low-density regions will have few samples and estimating the density ratio in these regions can be poor.
To address this, we leverage results in the normalizing flow literature, whereby the embedding space is transformed by an invertible map such that the resulting target and logging policies in the transformed embedding lie in a unit Gaussian ball \cite{choiFeaturizedDensityRatio2021}. This increases overlap probability and reduces finite sample pathological behavior in the density ratio estimation.

Let $n: \mathcal{G} \times {\cal C} \to \mathcal{Z}$ denote the normalizing flow, which is generally parameterized by a deep neural network whose overall architecture is such that $n^{-1}$ exists. 
Then, for $X \in {\cal G} \times {\cal C}$, its probability $\pi(X)$ can be evaluated exactly as
\[\pi(X) = t(n(X)) \abs{\det \frac{\partial n(X)}{\partial X}}\]
where the base distribution $t(\cdot)$ is generally chosen to be a multivariate standard normal distribution. First, $n(X) \sim N(0, I)$ which is a simple distribution.  Because $n$ is invertible, this transformation process does not lose information and the transformed embedding $n$ is of the same dimension as the original embedding $g$. Second, since data from both the logging and target policies are involved, the distance between these densities in the transformed embedded space is reduced.



We propose the following estimation strategy: First, we construct the tensor product embedding space $G \otimes C$, which represents a general composition of the embedded action and context spaces. Then, we fit the normalizing flow model ${n}$ given logging data ${\cal D}_0 = (G_i \otimes C_i)_{i=1}^N$, and target data ${\cal D}_1 = (G_i' \otimes C_i)_{i=1}^N$. Finally, we fit the probabilistic classifier $\hat{\eta}$ using features ${\cal D}_n = [{\cal D}_0, {\cal D}_1]$ and target $Z = 0$ for logging samples and $Z=1$ for target samples. Then, obtain weights $(\hat{w} \circ n)(G, C) = (\hat{\eta} \circ n) (G, C) / ( 1- (\hat{\eta} \circ n) (G, C))$, which represents the result of using a probabilistic binary classifier to learn the density ratio on the $n$-transformed space.



The \emph{featurized embedded permutation weighting self-normalized} estimator (FEPWS) is given as 
\begin{equation}
\hat{V}_{FEPWS} = \frac{\sum_{i=1}^N Y_i \hat{w}_{n}(G_i, C_i) K\left(\frac{G'_i - G_i)}{h}\right) }{\sum_{i=1}^N \hat{w}_{n}(G_i, C_i) K\left(\frac{G'_i - G_i}{h}\right)}\label{eqn:embedding-ope}.
\end{equation}

We show in \cref{sec:analysis} that this procedure is consistent.

\section{Theoretical Analysis}\label{sec:analysis}
In this section, we study the unbiasedness of the embedded and featurized permutation weighting estimators (EPW, FEPW) under both action space sizes and sample sizes. All proofs are deferred to the appendix.

As in \cite{sondhiBalancedOffPolicyEvaluation2020}, we assume certain conditions of the classifier:
\begin{assumption}\label{ass:classifier_error}
The classifier $\hat{\eta}$ must have error $\hat{\eta} - \eta$ scaling $O(N^{\epsilon})$ for  $\epsilon \in (0, 1)$; and must use a twice differentiable strictly proper scoring rule as its loss function.
\end{assumption}

\subsection{Analysis assuming positivity of embedding}

We first show that under our assumptions, the EPW estimator is unbiased and consistent. 
The regret under loss function $\lambda$ is defined as the difference between the risk of a classifier and the Bayes-optimal risk, $\textrm{reg}(\hat{\eta}; {\cal D}, \lambda) = \mathbb{L}(\hat{\eta}; {\cal D}, \lambda) -  \min_{\hat{\eta}} \mathbb{L}(\hat{\eta}; {\cal D}, \lambda)$, and under \cref{ass:classifier_error} this goes to zero as $N \to \infty$. Many loss functions (logistic, exponential, squared) also satisfy the strictly proper scoring rule property, and each such rule is associated with a Bregman divergence $h$.


\begin{restatable}{lem}{biasEpw}
\label{lem:bias_epw} (Bias of EPW)
Let $\E_0$ and $\E_1$ denote the expectations under the logging and target probability distributions $p_0(Y, G, C), p_1 (Y, A, C)$ respectively. Let $w(A, C)$ and $w(G, C)$ denote the 
density ratio under the action and embedding spaces respectively. Let ${\cal D} = [(G_i, C_i)_{i=1}^N, (G'_j, C_j)_{j=1}^N]$ denote the embedding and context data from both logging and target policies. Then,
\[\abs{\E_1[Y] - \E_0[Y \hat{w}(G, C)]} \leq \E[\abs{Y} \frac{2}{\sqrt{h''(1)}} \sqrt{\textrm{reg}(\hat{\eta}; {\cal D})}]\]
\end{restatable}

\begin{restatable}{lem}{varEpw}
\label{lem:var_epw}
(Variance of EPW) Let $w=w(G, C)$  and let $\Var_0 [Y \hat{w}(G, C)]$ denote the variance of the EPW estimator. Let ${\cal D} = [(G_i, C_i)_{i=1}^N, (G'_j, C_j)_{j=1}^N]$ denote the embedding and context data from both logging and target policies. Then
\begin{align*}
   &\Var_0[Y\hat{w}(G, C)] \\
   &\leq \frac{1}{N} \E_0[Y^2w(G, C)^2] + \frac{4}{N\sqrt{h''(1)}} \sqrt{\textrm{reg}(\hat{\eta}; {\cal D})} \E_0\left[Y^2 w_n(G, C) + \frac{1}{\sqrt{h''(1)}} \sqrt{\textrm{reg}(\hat{\eta}; {\cal D})}\right] 
\end{align*}
\end{restatable}

\begin{restatable}{lem}{consistencyEpw}
\label{lem:consistency_epw}(Consistency of EPW)
Let $\E_0$ and $\E_1$ denote the expectations under the logging and target probability distributions $p_0(Y, G, C) , p_1 (Y, A, C)$ respectively. Under \cref{ass:positivity,ass:consistency,ass:conditional_ig,ass:map,ass:classifier_error}, EPW is consistent -- that is, as $N \to \infty$, $\E_0[Y\hat{w}(G,C)] \to \E_1[Y]$.

\end{restatable}

We next establish that featurizing space of weights does not affect properties of asymptotic unbiasedness and consistency of the FEPWS estimator. 


\begin{restatable}{lem}{biasFepw} \label{lem:bias_fepw}

(Bias of FEPW)
Let $\E_0$ and $\E_1$ denote the expectations under the logging and target policies respectively. Let $w = w(G, C)$ and $w_n = (w\circ n)(G, C))$ denote the weight under the original and featurized embedding spaces respectively. Then,
\[\abs{\E_1[Y w] - \E_0[Y \hat{w}_n]} \leq \E[\abs{Y} \frac{2}{\sqrt{h''(1)}} \sqrt{\textrm{regret}(\hat{\eta})}]\]
\end{restatable}

\begin{restatable}{lem}{varFepw}\label{lem:var_fepw}
(Variance of FEPW)
Let $\Var_0[Y\hat{w}_n] $ denote the variance of the FEPW estimator. Then,
\[\Var_0[Y\hat{w}_n] \leq \frac{1}{N} \E_0[Y^2w^2] + \frac{4}{\sqrt{h''(1)}} \sqrt{\textrm{reg}(\hat{\eta})} \E_0\left[Y^2 w_n + \frac{1}{\sqrt{h''(1)}} \sqrt{\textrm{reg}(\hat{\eta})}\right]\]

\end{restatable}

\begin{restatable}{lem}{consistencyFepw}\label{lem:consistency_fepw}(Consistency of FEPW)
Let $\E_0$ and $\E_1$ denote the expectations under the logging and target probability distributions $p_0(Y, G, C) , p_1 (Y, A, C)$ respectively. Under \cref{ass:positivity,ass:consistency,ass:conditional_ig,ass:map,ass:classifier_error}, FEPW is consistent -- that is, as $N \to \infty$, $\E_0[Y\hat{w}_n(G,C)] \to \E_1[Y]$.
\end{restatable}
\subsection{Analysis without assuming positivity of embedding}
In this section we describe a situation where we do not have positivity over the action space, but also have a deterministic embedding. This means that both the policy $\pi(A \mid C)$ and the embedding $p(G \mid A)$ may be zero at certain places, and this can induce a positivity violation in $\pi(G \mid C) \sum_A p(G \mid A) \pi(A \mid C)$.
We argue that in fact this situation is likely to arise in practical usage of this algorithm. One motivation to consider embeddings in the first place is that positivity violations are occurring in the action space. Furthermore, embeddings obtained from deep neural networks tend to be deterministic, as they map each action to a particular embedding vector without randomness. 

The EPW(S) and FEPW(S) methods continue to work in this setting, since they rely on fitting a classifier to estimate the density ratio, and the classifier will impose smoothing assumptions per the choice of loss function. The key point is that at any given number of observed actions in the logging and target spaces, ($d_0 = \abs{{\cal A}_0}$ and $d_1 = \abs{{\cal A}_1}$ respectively),  some amount of bias will be incurred, regardless of the sample size $N$. 

For some $d_0, d_1$, as $N \to \infty$,
\[w^{obs}(G, C) = \frac{\pi_1^{obs}(G \mid C))}{\pi_0^{obs}(G \mid C)} \to \frac{\tilde{\pi}_1 (G \mid C))}{\tilde{\pi}_0(G \mid C)} = \tilde{w}(G, C).\]

Note that $\tilde{w} = \frac{\tilde{\pi}_1}{\tilde{\pi}_0} \neq \frac{\pi_1}{\pi_0} = w$, in general. The reason is that we must extrapolate from an observed policy $\pi^{obs}(G \mid C)$ over a finite set of observations (in $d_0$ or $d_1$) to the underlying policy $\pi(G \mid C)$ over a continuous embedding space. We denote this extrapolation as $\tilde{\pi}(G \mid C)$, and the distribution $\tilde{p}$ is defined as $p(Y \mid G, C)\pi(G \mid C) p(C)$.
To analyze the impact of the difference between the extrapolated policy $\tilde{\pi}$ and the true underlying policy $\pi$, we use the KL-divergence $\infdiv{\tilde{\pi}}{\pi}$ as a measure of the discrepancy. Furthermore we assume that this discrepancy decreases as we observe more items, such that 
\[d \to \infty \implies \infdiv{\tilde{\pi}}{\pi} \to 0.\]

We next introduce the quadratic cost transportation inequality to link the KL-divergence of two distributions, to the difference in expectations of a random variable under each of those distributions.
\begin{lem}\label{lem:quad_cost}
(Quadratic cost transportation inequality, \cite{boucheronConcentrationInequalitiesNonasymptotic2013a})
Let $Z$ be a real-valued integrable random variable. Then, given $\nu > 0$
\[\log \E[\exp(\lambda (Z - \E[Z]))] \leq \frac{\nu \lambda^2 }{2}\]

for every $\lambda > 0$ if and only if for any probability measure $Q$ absolutely continuous with respect to $P$ such that $\infdiv{Q}{P} < \infty$, 
\[\E_Q [Z] - \E_P[Z] \leq \sqrt{2 \nu \infdiv{Q}{P}}\]
\end{lem}

Lemma \ref{lem:quad_cost} requires that the moment generating function is bounded above by a quadratic function. This admits a wide class of parametric distributions (e.g. Gaussians). As such for the purposes of further analysis we make the following assumption:
\begin{assumption} \label{ass:quad_cost}
There exists  $\nu_0 >0$, such that for all $\lambda_0 > 0$,
\[\log \E[\exp(\lambda (Yw_{obs}  - \E_{\tilde{p}_0}[Y w_{obs}]))] \leq \frac{\nu_0 \lambda_0^2}{2},\]
and there exists  $\nu_1 > 0$ such that  for all $\lambda_1 > 0$
\[\log \E[\exp(\lambda ((Yw_{obs})^2 - \E_{\tilde{p}_0}[(Y w_{obs})^2]))] \leq \frac{\nu_1 \lambda_1^2}{2},\]
\end{assumption}

Then given Assumption \ref{ass:quad_cost} and Lemma \ref{lem:quad_cost}, we can derive the following bounds utilizing the quadratic cost transportation inequality, a strong form of the Gaussian concentration property~\citep{cattiaux2006quadratic}:

\begin{restatable}{cor}{quadCost}\label{cor:quad_cost}
(Applying the quadratic cost transportation inequality)
Under Assumption \ref{ass:quad_cost}, we have that 
\[\E_{\tilde{p}_0} [Y w_{obs}] - \E_{p_0}[Y w_{obs}] \leq \sqrt{2 \nu_0 \infdiv{\tilde{\pi}_0 }{\pi_0}}\]
 and 
\[\E_{\tilde{p}_0} [(Y w_{obs})^2] - \E_{p_0}[(Y w_{obs})^2] \leq \sqrt{2 \nu_1 \infdiv{\tilde{\pi}_0 }{\pi_0}}\]
\end{restatable}

We can then apply these results to analyse the error of the density ratio based estimator of \cite{sondhiBalancedOffPolicyEvaluation2020}, using techniques from \cite{arbourPermutationWeighting2021}.


\begin{restatable}{lem}{biasVarEpw}
Let $\kappa_r = \sqrt{\textrm{reg}(\hat{\eta}; \mathcal{D}})$ denote the regret of the classifier, where $\textrm{reg}(\hat{\eta}; \mathcal{D}) = \mathbb{L}(\hat{\eta}; \mathcal{D}, \gamma) - \min_{\hat{\eta}} \mathbb{L}(\hat{\eta}; \mathcal{D}, \gamma)$ and let $h$ denote some Bregman generator. Let $\mathrm{Bias} = \E_{\tilde{p}_0} [Y w^{obs}] - \E_{p_0} [Y w]$. Then, 
\begin{align*}
    \mathrm{Bias} 
    & \leq \underbrace{\E_{{p}_0} \left[\frac{2\abs{Y} }{\sqrt{h''(1)}} \kappa_r\right]}_{\text{error due to sample size}} + \underbrace{\sqrt{2 \nu_0 \infdiv{\tilde{\pi}_0}{\pi_0}}}_{\text{error due to irreducible discrepancy of $\pi_0$ and $\tilde{\pi}_0$}},\\
\Var_{\tilde{p}_0}[Y w^{obs}] 
        &\leq \underbrace{\E_{p_0}[Y^2 (w^{obs})^2] + \E_{p_0} \left[Y^2 \left(\frac{4w}{\sqrt{h''(1)}} {\kappa_r}  \frac{4}{h''(1)}\kappa_r^2\right)\right]}_{\text{variance bound due to sample size}} + \\
        &+\underbrace{\sqrt{2 \nu_1 \infdiv{\tilde{\pi}_0}{\pi_0}}}_{\textrm{variance bound due to irreducible discrepancy of $\pi_0$ and $\tilde{\pi}_0$} }.
\end{align*}
\end{restatable}

\section{Experiments} \label{sec:results}

Performance is measured using root mean squared error. For $s$ indexing $S$ simulations, $\textrm{RMSE} = \sqrt{\frac{1}{S} \sum_{s=1}^S (\hat{V}_s - V_s) ^ 2}$
where $V_s= \frac{1}{N} \sum_{i=1}^N Y'_{i, s}$ is the observed reward under the target policy in simulation trial $s$.

All details of experiments are in the appendix.
\subsection{Toy Simulation } \label{sec:sim_setup}

\subsubsection{Estimators where weights are given}
We first consider scenarios in which estimators (DM, IPW, EIPW) have access to policy density functions. The DM estimator uses the \emph{scikit-learn} implementation of \texttt{GradientBoostedRegressor}, while the IPW and EIPW estimators do not have models to fit (since in this scenario oracle weights are given). We first consider a fixed sample size of $N=1000$ and number of actions $\abs{\cal A} \in \{10, 50, 100, 200, 500\}$. We observe that the performance of the embedded system improves on the IPW and DM estimators, especially at higher number of actions. 

Similar trends are observed when we fix the number of actions to $\abs{\cal A} = 100$ and vary the sample sizes $N \in \{100, 500, 1000, 2000, 5000\}$. While all estimators improve in performance with increasing sample size, EIPW retains its position as the best performing estimator. 

\begin{figure}[t]
     \centering
     \begin{subfigure}[b]{0.23\textwidth}
         \centering
         \includegraphics[width=\textwidth]{"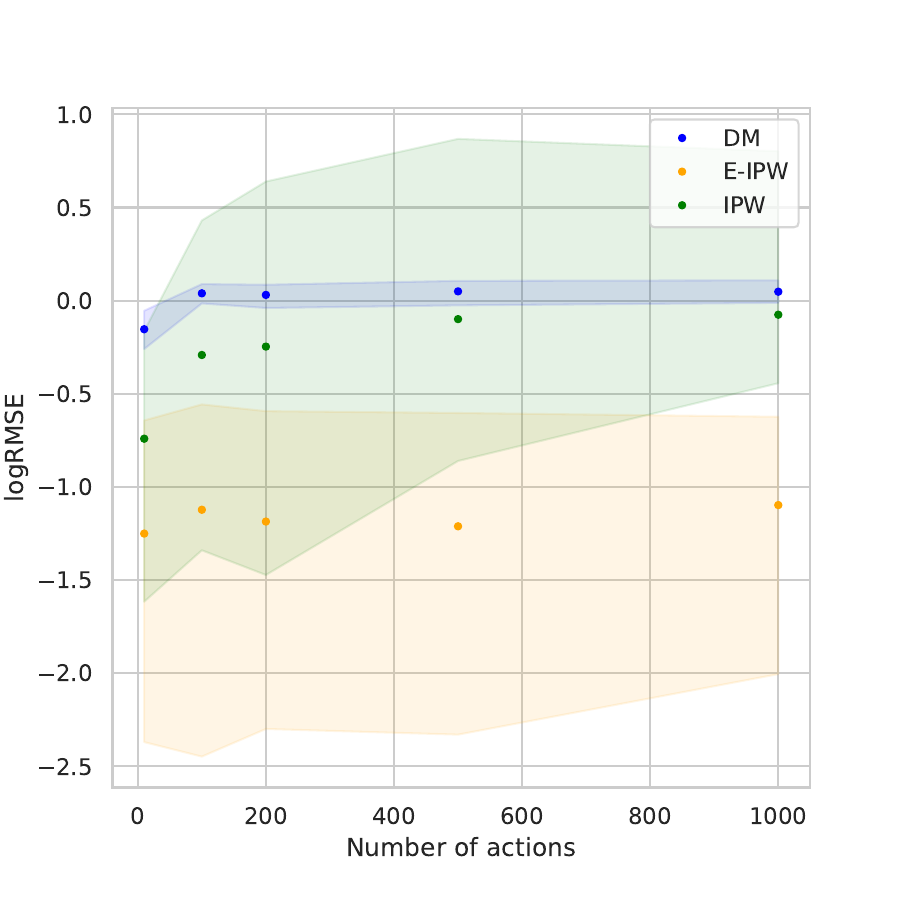"}
         \caption{}
         \label{fig:n_actions}
     \end{subfigure}
     \begin{subfigure}[b]{0.23\textwidth}
         \centering
         \includegraphics[width=\textwidth]{"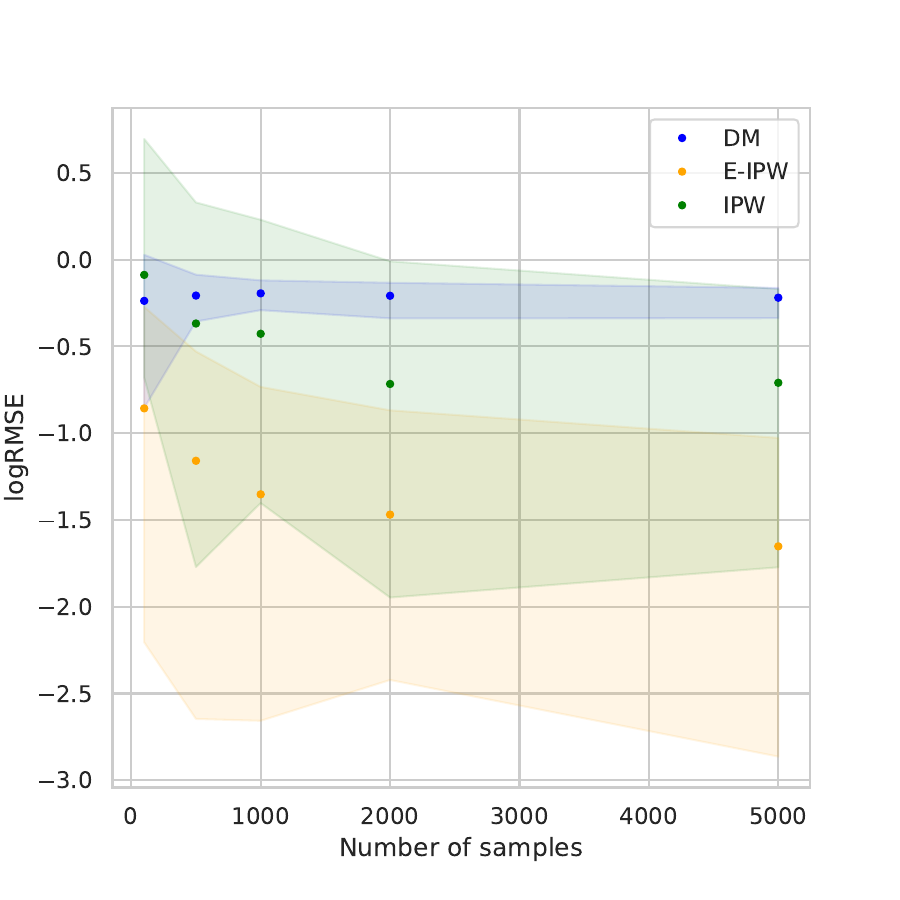"}
         \caption{}
         \label{fig:n_samples}
     \end{subfigure}
     \begin{subfigure}[b]{0.23\textwidth}
         \centering
         \includegraphics[width=\textwidth]{"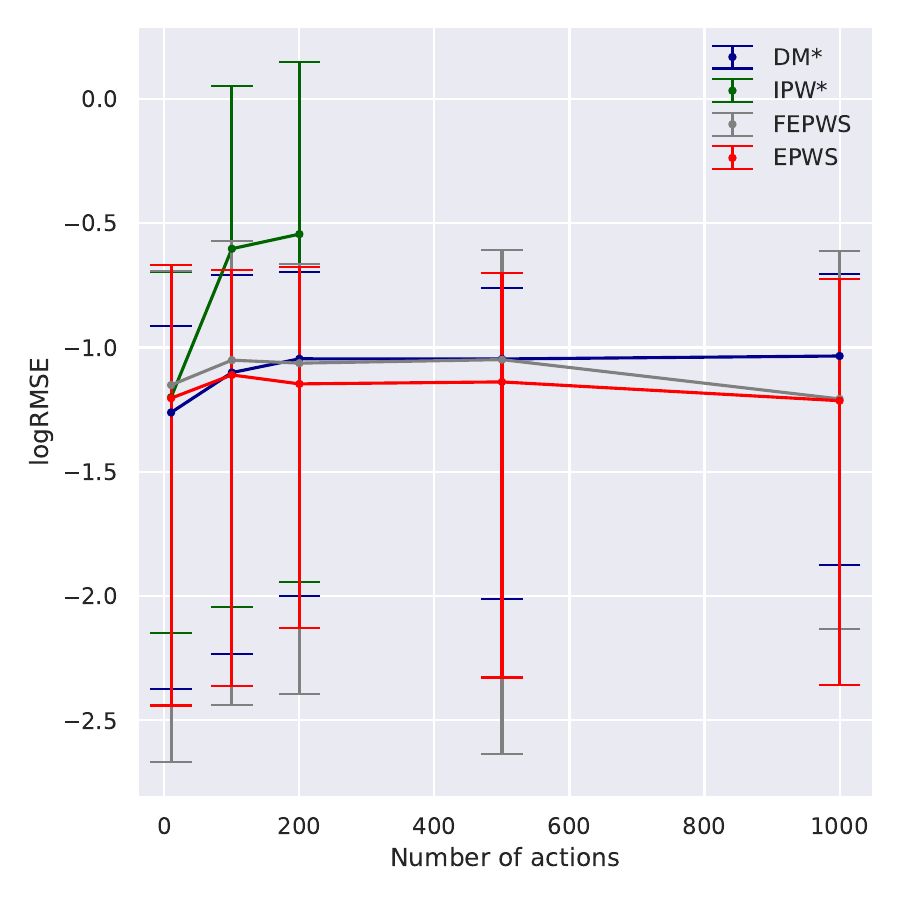"}
         \caption{}
         \label{fig:rmse_failures}
     \end{subfigure}
     \begin{subfigure}[b]{0.23\textwidth}
         \centering
         \includegraphics[width=\textwidth]{"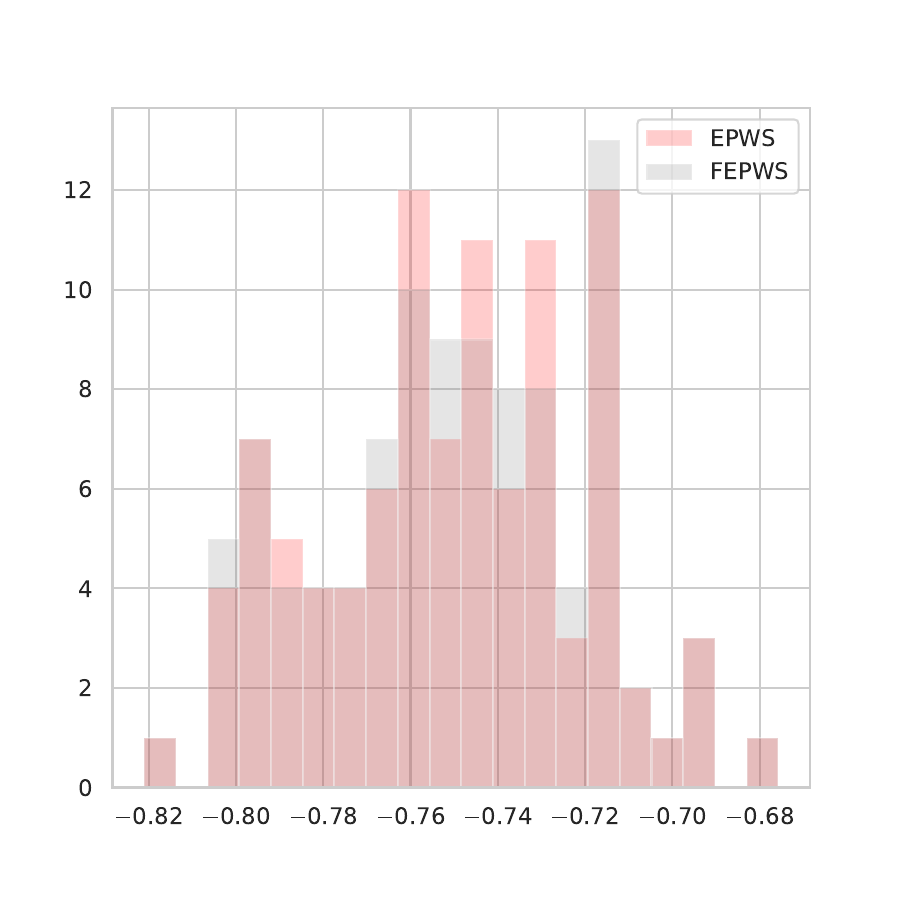"}
         \caption{}
         \label{fig:realistic_experiment}
     \end{subfigure}
     \caption{\cref{fig:n_actions,fig:n_samples} detail $\log_{10} \textrm{RMSE}$ performance in increasing number of actions and samples respectively, when estimators have the true policy densities; 
     \ref{fig:rmse_failures} shows how IPW* (which estimates the density ratio over the action space ${\cal A}$ fails to perform completely above 500 actions; \ref{fig:realistic_experiment} shows performance of FEPWS and EPWS on the Behance dataset.
     }
\end{figure}



\subsubsection{Practical advantages of embedded estimators}
To understand the problem better we introduce some simple simulation studies aimed at studying properties of the estimators. Since the purpose of these estimators is to function where standard off-policy estimators fail, a direct comparison of error is not possible. Instead, we compare the robustness of these embedded estimators  (FEPWS, EPWS) against standard estimators (IPW*, DM*), in scenarios where all estimators must estimate their weights from data.

We consider ${\cal A} \in \{10, 100, 200, 500, 1000\}$ with $N=500$ and 100 simulated datsets. The results show that IPW* fails to perform at around 500 actions, due to the positivity violations encountered. While the direct method continues to perform, it is also more susceptible to model misspecification. By contrast, FEPWS and EPWS are both resistant to these issues, and have relatively good performance across the range of action sizes.

\subsection{Realistic Semisynthetic Experiment}
We apply our method to a semi-synthetic simulation based on the publicly available Behance dataset from \cite{heVistaVisuallySocially2016}. The dataset contains a selection of 1 million appreciates ("likes") on 178,788 items from 63,497 users. A portion of these users are content creators, in that they have contributed to creating at least one item. Furthermore, each item has a 4096-dimensional embedding extracted from a VGG neural network. We consider context $C$ to be a binary variable indicating if a user is a content creator; action $A$ to be a particular item, embedding $G$ to be the corresponding neural network embedding, and reward $Y$ to be appreciates.

Both estimators perform well in this context, demonstrating the viability of embedding methods on real-world data. Due to computational constraints only a compressed version of the embedding was used; we hypothesize that performance would increase with a higher quality embedding. 

\section{Conclusion} \label{sec:conc}
In this paper we propose an embedding space approach to the IPW that leverages normalizing flows. We demonstrate theoretical properties of this estimator under random and deterministic embeddings. Finally we provide experiments on a toy example as well as a semi-synthetic application.


\begin{ack}
Use unnumbered first level headings for the acknowledgments. All acknowledgments
go at the end of the paper before the list of references. Moreover, you are required to declare
funding (financial activities supporting the submitted work) and competing interests (related financial activities outside the submitted work).
More information about this disclosure can be found at: \url{https://neurips.cc/Conferences/2022/PaperInformation/FundingDisclosure}.

Do {\bf not} include this section in the anonymized submission, only in the final paper. You can use the \texttt{ack} environment provided in the style file to autmoatically hide this section in the anonymized submission.
\end{ack}

\section*{Checklist}

\begin{enumerate}

\item For all authors...
\begin{enumerate}
  \item Do the main claims made in the abstract and introduction accurately reflect the paper's contributions and scope?
    \answerYes{}
  \item Did you describe the limitations of your work?
    \answerYes{}
  \item Did you discuss any potential negative societal impacts of your work?
    \answerNA{}
  \item Have you read the ethics review guidelines and ensured that your paper conforms to them?
    \answerYes{}
\end{enumerate}

\item If you are including theoretical results...
\begin{enumerate}
  \item Did you state the full set of assumptions of all theoretical results?
    \answerYes{}
        \item Did you include complete proofs of all theoretical results?
    \answerYes{}
\end{enumerate}

\item If you ran experiments...
\begin{enumerate}
  \item Did you include the code, data, and instructions needed to reproduce the main experimental results (either in the supplemental material or as a URL)?
    \answerYes{}
  \item Did you specify all the training details (e.g., data splits, hyperparameters, how they were chosen)?
    \answerYes{}
        \item Did you report error bars (e.g., with respect to the random seed after running experiments multiple times)?
    \answerYes{}
        \item Did you include the total amount of compute and the type of resources used (e.g., type of GPUs, internal cluster, or cloud provider)?
    \answerYes{}
\end{enumerate}

\item If you are using existing assets (e.g., code, data, models) or curating/releasing new assets...
\begin{enumerate}
  \item If your work uses existing assets, did you cite the creators?
    \answerYes{}
  \item Did you mention the license of the assets?
    \answerNA{}
  \item Did you include any new assets either in the supplemental material or as a URL?
    \answerYes{}
  \item Did you discuss whether and how consent was obtained from people whose data you're using/curating?
    \answerNA{}
  \item Did you discuss whether the data you are using/curating contains personally identifiable information or offensive content?
    \answerNA{}
\end{enumerate}

\item If you used crowdsourcing or conducted research with human subjects...
\begin{enumerate}
  \item Did you include the full text of instructions given to participants and screenshots, if applicable?
    \answerNA{}
  \item Did you describe any potential participant risks, with links to Institutional Review Board (IRB) approvals, if applicable?
    \answerNA{}
  \item Did you include the estimated hourly wage paid to participants and the total amount spent on participant compensation?
    \answerNA{}
\end{enumerate}

\end{enumerate}

\bibliographystyle{unsrtnat}
\bibliography{references,library}
\newpage
\appendix
{\huge \bf Appendix}

\section{Proofs}

\biasEpw*
\begin{prf}
\begin{align}\label{eqn:sum_terms}
    \E_0 [Y w(G, C) ] & =  \E_0 [Y w(G, C)] + \E_0 [Y (\hat{w} (G, C) - w (G, C))]\\
\end{align}
where the third equality follows by application of \cref{ass:map}

Inspecting the first term of \cref{eqn:sum_terms}, we find that 
\begin{align*}
    \E_0 [Y w(G, C)] &= \sum_{YGC} Y w(G, C) p_0(Y, G, C) \\
    &= \sum_{YGC} Y \frac{\pi_1(G \mid C)}{\pi_0 (G \mid C)} p(Y \mid G, C) p_0 (G \mid C) p(C) \\
    &= \sum_{YGCA} Y (\sum_{A} p(G \mid A)\pi_1 (A \mid C)) p(Y \mid G, C) p(C) \\
    &= \sum_{YGCA} Y p(Y \mid G, C) p(G \mid A) \pi_1 (A \mid C) p(C) \\
    &= \sum_{YCA} Y p(Y \mid A, C) \pi_1 (A \mid C) p(C)\\
    &= \E_1 [Y]
\end{align*}
where the third equality applies \cref{ass:map} and cancels $\pi_0 (G \mid C)$, and the fifth equality applies \cref{ass:map} again. 

Then, we apply Lemma A.1 of \cite{arbourPermutationWeighting2021}, which allows us to bound the last term of \cref{eqn:sum_terms}, such that
\begin{align*}
    \E_0 [Y (\hat{w}(G, C) - w(G, C))] & \leq \E_0[\abs{Y}\abs{\hat{w}(G, C) - w(G, C)}] \\
    &\leq \E_0 \left[\abs{Y} \abs{\frac{2}{\sqrt{h''(1)}} \sqrt{\textrm{reg}(\hat{\eta})}} \right]
\end{align*}

\end{prf}


\varEpw*
\begin{prf}

The result follows from applying Proposition 4.2 of \cite{arbourPermutationWeighting2021}.

\end{prf}

\consistencyEpw*
\begin{prf}
Since the regret of the classifier goes to zero as $N \to \infty$, with bounded $Y$ the bias and variance of EPW goes to zero.
\end{prf}

We recall an important result from \cite{choiFeaturizedDensityRatio2021}.

\begin{lem}\label{lem:f_dre_equiv} (Lemma 1, \cite{choiFeaturizedDensityRatio2021})
Let $X_p \sim p$ be a random variable with density $p$, and $X_q \sim q$ a random variable with density $q$. Let $p', q'$ be densities of $n(X_p
), n(X_q)$ respectively. Let $n(\cdot)$ be an invertible mapping. Then for any value $x$
\[ \frac{p(x)}{q(x)} = \frac{p'(n(x))}{q'(n(x))}\]
\end{lem}
\cref{lem:f_dre_equiv} states that the true weights remain unchanged under a valid normalizing flow $n(\cdot)$. 

\biasFepw*

\begin{prf}

Note that
\begin{align}\label{eqn:sum_2}
    \E_0 [Y\hat{w}_n (G, C)]  &= \E_0 [Y w_n (G, C)] + \E_0[Y(\hat{w}_n (G, C) - w_n (G, C)] 
\end{align}

However, by \cref{lem:f_dre_equiv} it is established that $w(G, C) = w_n(G, C)$, and therefore $\E_0[Y w_n(G, C)] =\E_0[Y w(G, C)] $. 
The rest of the proof involving the bounding of the second term of \cref{eqn:sum_2} follows the argument in \cref{lem:bias_epw}.
\end{prf}

\varFepw*
\begin{prf}
Following \cite{arbourPermutationWeighting2021}, we note that the second moment serves as an upper bound for the variance. Then,

\begin{align} \label{eqn:sum_3}
   \E_0 [(Y \hat{w}_n(G, C))]  &= \E_0 [Y^2 w^2_n] + \E_0[Y^2 (2 w_n (\hat{w}_n - w_n) + (\hat{w}_n - w_n)^2) ] 
\end{align}

The first term of \cref{eqn:sum_3} can be expressed as $\E_0 [Y^2 w^2]$ through application of \cref{lem:f_dre_equiv}. The second term of \cref{eqn:sum_3} follows from Lemma A.1 of \cite{arbourPermutationWeighting2021}.

\end{prf}

\consistencyFepw*
\begin{prf}
Since the regret of the classifier goes to zero as $N \to \infty$, with bounded $Y$ the bias and variance of FEPW goes to zero.
\end{prf}

\biasVarEpw*
\begin{prf}
\begin{align*}
    \mathrm{Bias} & = \E_{\tilde{p}_0}[Y w^{obs}] - \E_{p_0} [Y w], \\
    & \leq \E_{{p}_0}[Y w^{obs}] - \E_{p_0} [Y w] + \sqrt{2 \nu_0 \infdiv{\tilde{\pi}_0}{\pi_0}},\\
    & \leq \underbrace{\E_{{p}_0} \left[\frac{2\abs{Y} }{\sqrt{h''(1)}} \kappa_r\right]}_{\text{error due to sample size}} + \underbrace{\sqrt{2 \nu_0 \infdiv{\tilde{\pi}_0}{\pi_0}}}_{\text{error due to irreducible discrepancy of $\pi_0$ and $\tilde{\pi}_0$}},\\
    \end{align*}

where the first line states the definition of the bias, the second line applies the result of Corollary \ref{cor:quad_cost}, and the third line applies the result of Proposition 4.1 in \cite{arbourPermutationWeighting2021}.

The bound on the bias has two terms. The first term is due to the loss incurred due to the Bregman generator of the classifier, which vanishes as $N \to \infty$. The second term is due to the discrepancy between the underlying policy $\pi$ and the extrapolated policy $\tilde{\pi}$, which vanishes as $d \to \infty$.
    
We then compute an upper bound for the variance of the estimator. 
\begin{align*}
&\Var_{\tilde{p}_0}[Y w^{obs}] \\
    &\leq \E_{\tilde{p}_0} [(Y w^{obs})^2] \\
    &\leq \E_{p_0} [(Y w^{obs})^2] + \sqrt{2 \nu_1 \infdiv{\tilde{\pi}_0}{\pi_0}}\\
        &\leq \E_{p_0}[Y^2 (w^{obs})^2] + \E_{p_0} \left[Y^2 \left(\frac{4w}{\sqrt{h''(1)}} {\kappa_r} + \frac{4}{h''(1)}\kappa_r^2\right)\right] + \sqrt{2 \nu_1 \infdiv{\tilde{\pi}_0}{\pi_0}},  \\
\end{align*}
where the first inequality follows from the fact that the second moment is a trivial upper bound for the variance, the second from Corollary \ref{cor:quad_cost}, and the third by application of Proposition 4.2 from \cite{arbourPermutationWeighting2021}.
\end{prf}

\quadCost*
\begin{prf}
We apply Lemma \ref{lem:quad_cost} to the quantities $Y w_{obs}$ and $(Y w_{obs})^2$. Furthermore, we note that because $G$ is a deterministic function of $A$, a bijection can be established such that $p(Y \mid G, C) = p ( Y  \mid A, C)$.  Finally, by the chain rule property of KL-divergences,
\[\infdiv{\tilde{p}}{p} = \infdiv{p(Y \mid A, C)}{p(Y \mid A, C)} + \infdiv{\tilde{\pi}}{\pi} + \infdiv{p(C)}{p(C)} = \infdiv{\tilde{\pi}}{\pi} \]
\end{prf}

\section{Experiments}
In this section we detail experiments performed in the main paper. Experiments were performed using a Lenovo Legion 5 with 16 GB of RAM, an NVIDIA GeForce 3060 RTX graphics card, and an AMD Ryzen 5800 CPU.
\subsection{Toy Simulation Setup}

We adapt the toy simulation in \cite{saitoOffPolicyEvaluationLarge2022}.

The context $C$ is sampled from normal distribution with dimension $d_c = 2$. The action $A$ is sampled from a finite set ${\cal A}$. We define the embedding function as 
\[p(G | A) = \prod_{k=1}^{d_G} \frac{\exp{\alpha_{A, G_k}}}{\sum_{G' \in {\cal G}_k}\exp{\alpha_{A, G'}}}\]
where $d_G = 2$ is the dimension of the embedding, each dimension ${\cal G}_k$ has cardinality $\mathfrak{c}_G = 2$, where parameters $\alpha_{A, G_k    }$ are sampled i.i.d. from standard normal distribution. Each embedding vector $G \in {\cal G}$ is mapped to a context vector $x_G \in \mathbb{R}^d_G$, which is a randomly sampled vector from a standard normal distribution. The reward function is a function of $G$ only through $x_G$.
        
The reward (as a function of embedding and context) is given by the function 
\[Y(G=g, C=c) = \sum_{k=1}^{d_g} \eta_k (c^T B x_g + \beta_C^T c + \beta_G^T x_g)\]
where coefficient vectors $\beta_C, \beta_G$ and matrix $B$ are all sampled from uniform in range $(-1, 1)$, and $\eta_k$ are drawn from Dirichlet distribution $\textrm{Dir}({\bm \alpha} = \mathbf{1})$. The counterfactual reward as a function of action and context is defined as $Y(A=a, C=c) = \E_{p(G | A=a)}[Y(g, c)] $.
        
The logging policy $\pi_0$ is exponential tilted version $Y(a, c)$:
\[\pi_0 (A \mid C) = \frac{\exp(\beta_0 Y(a, c))}{\int_A \exp(\beta_0 Y(a, c))}\] 
where we choose $\beta_0 =-1$. The target policy is an $\epsilon$-deviation from the optimal policy maximizing $Y(a, c)$:
\[\pi_1(A \mid C) = (1 - \epsilon) I(A = \arg \max_{a' \in {\cal A}} Y(a', c)) + \epsilon / \abs{\cal A}\]        where we choose $\epsilon = 0.1$

For each of $N$ rows of the logging policy dataset, we sample a context $C_i$, which then allows us to sample action $A_i$ from the logging policy. Then, $G_i$ is sampled based on action $A_i$. Finally, the reward $Y_i$ is sampled based on the $G_i, C_i$. We generate the target policy dataset in the same way, except that the rewards are not provided to estimators (and used only for computing losses).

\subsubsection{Estimators where weights are estimated}
We now consider scenarios in which only data from the logging and target policies are available. We implement the EPWS and FEPWS estimators. For the estimation of the permutation weights, both estimators rely on \texttt{GradientBoostedClassifier}. The EIPW estimator is included as comparison for a hypothetical best weight estimation strategy using the initial embedding.

The PyTorch normalizing flow implementation is adapted from \cite{choiFeaturizedDensityRatio2021}, which in turn adapts the Masked Autoregressive Flow model of \cite{papamakariosNormalizingFlowsProbabilistic2021}. 

\subsection{Realistic Experiment}

To construct the reward function from data, we compute for each action and context the following probability:
\[p(Y= 1 \mid A=a, C=c) = \epsilon_{m} * \frac{\textrm{\#(appreciates for item $a$ in users with context $c$})}{\max_{a' \in \mathcal{A}} \textrm{\#(appreciates for item $a'$ in users with context $c$})}\]
where $\epsilon_m$ is a tuning parameter represents the maximum probability of an appreciate, which is set to $\epsilon_m=.1$. Note that in this example, the counterfactual reward $Y(a, c)$ is precisely the probability $p(Y=1 \mid A=a, C=c)$.

The embedding function maps actions to a particular continuous embedding vector -- thus, previous comments on bias incurred with deterministic embeddings apply. For computational tractability, we use \texttt{IncrementalPCA} from scikit-learn to compress the embedding down to 4 dimensions.

We employ the same logging and target policies as in \cref{sec:sim_setup}, except $\beta_0=0$.

For the simulation, we assume that $p(C=1) = 0.5$ since the value estimated from the data is $\hat{p}(C=1) \approx 0.003$.

We compute estimates for the FEPWS and EPWS estimators, with 100 bootstrap samples, at a sample size of $N=10000$. Given the large number of items, the probability that absolute continuity on the $A$ scale is violated is very high, but these estimators are able to circumvent this using the embedded space.

\section{Additional Experiments}
We conduct additional experiments where we increase the dimension of the context and embedding space to $d_C =d_G = 3$, and increase the number of bootstraps to 500.

\begin{figure}[t]
     \centering
     \begin{subfigure}[b]{0.5\textwidth}
         \centering
         \includegraphics[width=\textwidth]{"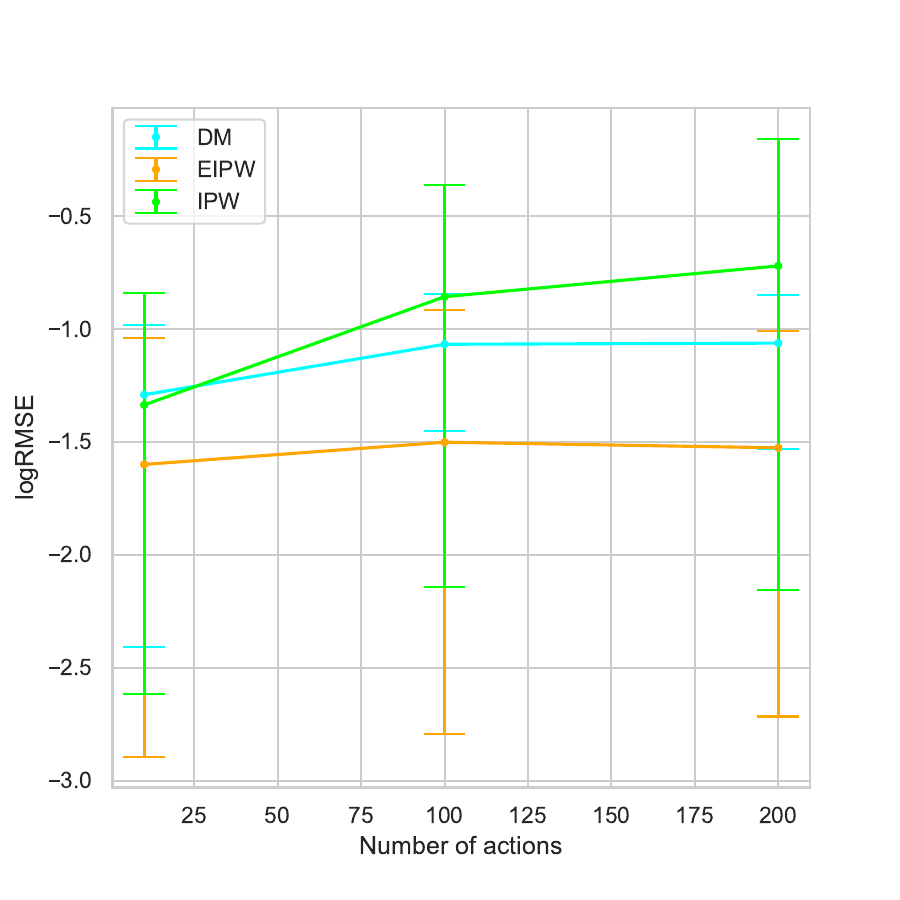"}
         \caption{}
         \label{fig:appendix_n_actions}
     \end{subfigure}
     
     \begin{subfigure}[b]{0.5\textwidth}
         \centering
         \includegraphics[width=\textwidth]{"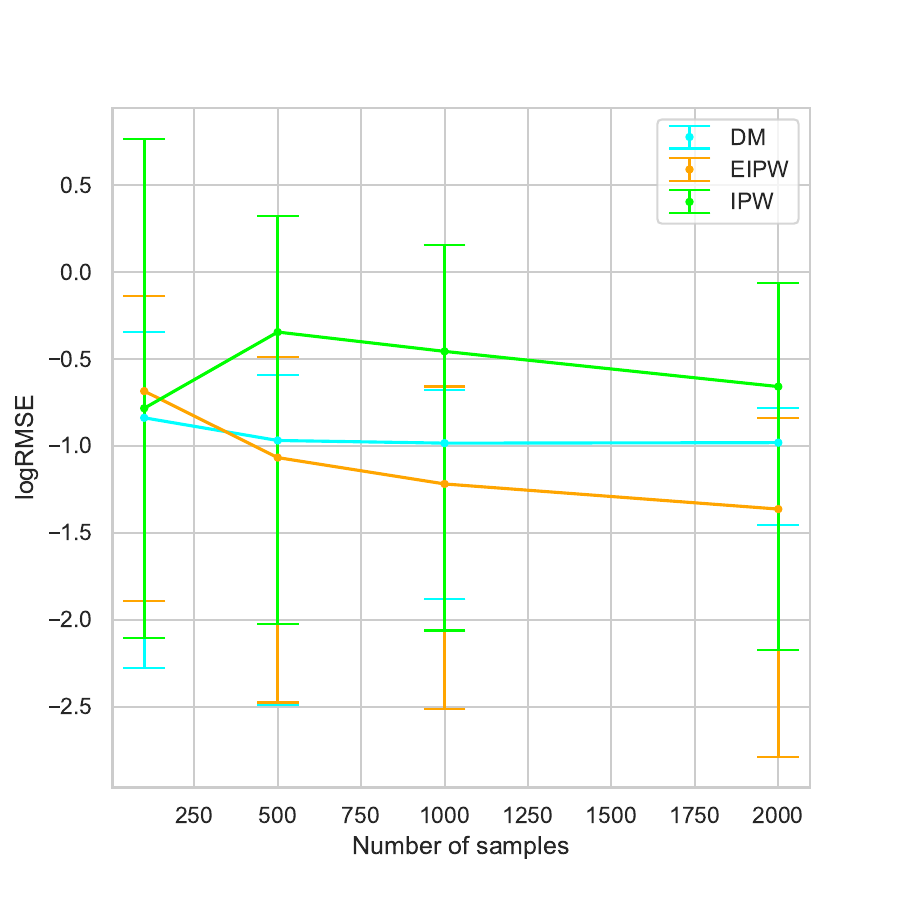"}
         \caption{}
         \label{fig:appendix_n_samples}
     \end{subfigure}

     \caption{\cref{fig:appendix_n_actions} covers additional experimentation in the number of actions, while \cref{fig:appendix_n_samples} covers additional experimentation in number of samples.}
     \label{fig:appendix_experiments}
\end{figure}

First, we consider $N=3000$ and ${\cal A} \in \{ 10, 100, 200\}$ in \cref{fig:appendix_n_actions}. Note that these results are conducted with estimators having access to the true weights. These results indicate that the performance of EIPW stays stable while the performance of the DM and IPW estimators worses with increasing number of actions. 

Second, we consider $N \in \{100, 500, 1000, 2000\}$ and ${\cal A} = 200$. While all estimators appear to benefit from increasing number of samples, the downward trend on the error for the EIPW estimator appears to be stronger than the DM or IPW estimators.
\end{document}